\documentclass[11pt]{article}

\usepackage[final]{acl}

\usepackage{times}
\usepackage{latexsym}

\usepackage[T1]{fontenc}

\usepackage[utf8]{inputenc}

\usepackage{microtype}

\usepackage{inconsolata}

\usepackage{graphicx}
\usepackage[table]{xcolor}
\usepackage{algorithm}
\usepackage{algorithmic}
\usepackage{booktabs}
\usepackage{multirow}
\usepackage{array}
\usepackage{amsmath} 
\usepackage{makecell}
\usepackage{graphicx}
\usepackage[export]{adjustbox}       

\title{GIFT: Goal-Injected Fine-Tuning for Efficient Manipulation Policy Adaptation}
\author{
 \textbf{Xiaoyuan Fang},
 \textbf{Shuo Feng},
 \textbf{Yuxuan Wang},
 \textbf{Enhua Cheng},
 \textbf{Peng Zhou},
 \textbf{Piji Li}\textsuperscript{$\dagger$}
\\
College of  Artificial Intelligence, \\
Nanjing University of Aeronautics and Astronautics, Nanjing, 211106, China\\
The Key Laboratory of Brain-Machine Intelligence Technology, \\ Ministry of Education, Nanjing, 211106, China.
\\
 \small{
   \textbf{E-mail:} \href{mailto:frankfang@nuaa.edu.cn,pjli@nuaa.edu.cn}
{\{frankfang,pjli\}@nuaa.edu.cn}
 }
}

\newcommand\blfootnote[1]{%
  \begingroup
  \renewcommand\thefootnote{}\footnote{#1}%
  \addtocounter{footnote}{-1}%
  \endgroup
}

\begin{document}
\maketitle
\blfootnote{$\dagger$ Corresponding author.}
\begin{abstract}

Compared with relying solely on initial observations and language instructions, predicting goal images with generative models as high-level visual guidance can significantly enhance the robustness of Vision-Language-Action (VLA) models. However, most existing foundation models have not systematically incorporated goal image conditioning due to the high computational training cost. To this end, we propose Goal-Injected Fine-Tuning (GIFT), a lightweight and efficient fine-tuning framework that seamlessly integrates generated goal images into multiple representative pretrained VLA models. Our approach introduces goal image features into observations via a zero-initialized convolution which progressively grows parameters from zero and prevents harmful noise from disrupting the pretrained policy during fine-tuning. As training proceeds, goal information is gradually incorporated, enabling efficient goal understanding without disrupting model stability. We further introduce a refined image editing method to generate semantically and visually consistent goal images from initial observations and task instructions. Experiments show that goal-aware VLA models achieve substantial performance gains across tasks: with only a single epoch of fine-tuning, GIFT outperforms the base model by 6.0\% and 13.4\% on two SIMPLER settings, and by 4.7\% on LIBERO, demonstrating both efficiency and effectiveness.

\end{abstract}

\section{Introduction}

\begin{figure}[t]
  \centering
  \includegraphics[width=\columnwidth]{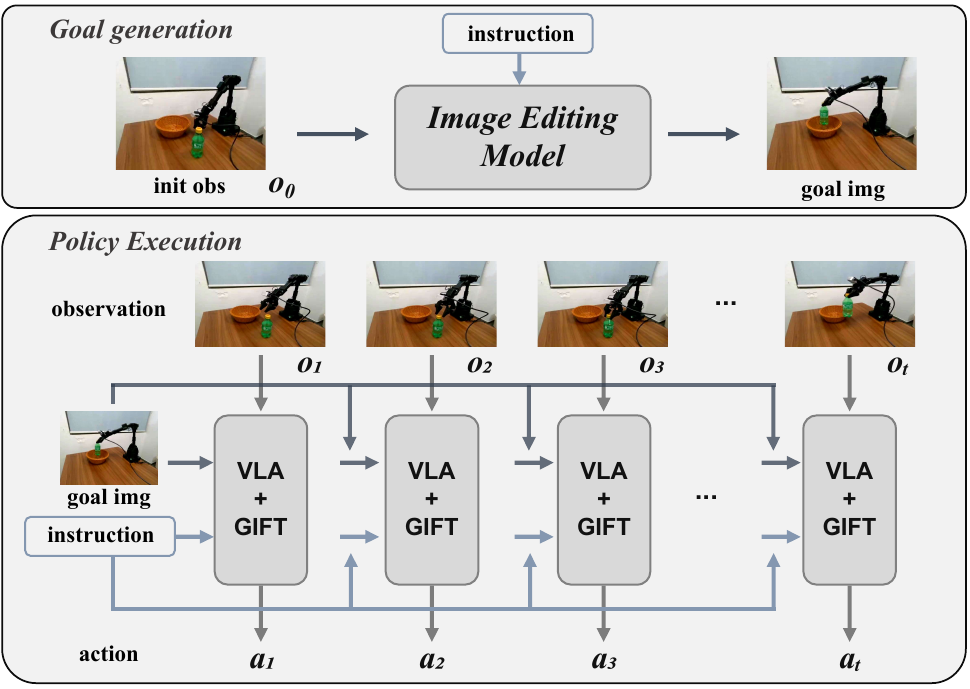}
  \vspace{-4pt}
  \includegraphics[width=\columnwidth]{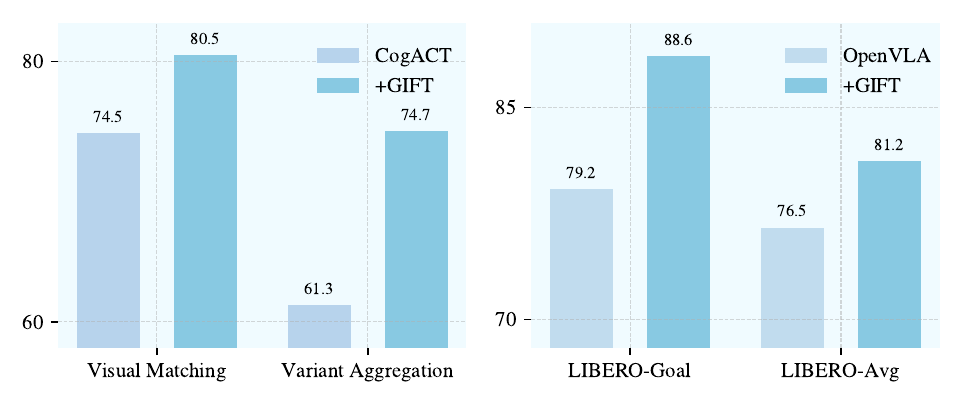} 
  \caption{
  Overview of the proposed GIFT framework.
  The image editing model first generates a goal image based on the initial observation and given instruction. The generated goal is then reused throughout policy execution to guide the VLA model at each timestep.
  }
  \label{fig1}
  \vspace{-8pt}
\end{figure}

Recent progress in robot learning has led to training strategies that perform robustly across diverse tasks and environments\cite{huang2026graphcot,kim2025fine,bi2025motus}. Vision-Language-Action (VLA) Models have emerged as a promising unified approach, using Vision-Language Models (VLMs) to map instructions and visual observations to actions \cite{intelligence2025pi_,bjorck2025gr00t}. By training VLMs on robot demonstrations, VLA models effectively transfer knowledge of scenes, objects and language to improve performance on manipulation tasks.

Despite progress in VLA models, their generalization is limited by scarce and heterogeneous robotic data, leading to degraded out-of-distribution performance. Recent works leverage generative and predictive modules pretrained on large-scale internet data for visual-semantic reasoning and goal generation~\cite{chen2025goal,zhou2025act2goal}. These models generate goal or future-state images to guide low-level policies, forming hierarchical designs analogous to human perception-to-action processes, which enhance manipulation performance~\cite{du2023learning,cheang2024gr} while complementing earlier generative subgoal approaches~\cite{blackzero,hatch2024ghil}.

Existing goal-conditioned VLA approaches face two main challenges. Some methods generate intermediate goal images for low-level policies but do not exploit pretrained VLA backbones~\cite{chen2025goal,zhou2025act2goal,hatch2024ghil,bharadhwajgen2act}, which limits their generalization to novel tasks. Other approaches integrate goal generation or predictive world modeling directly into the VLA pipeline, potentially altering pretrained reasoning pathways and often requiring substantial joint retraining~\cite{team2026motubrain,sun2026vla,bi2025motus,zhao2025cot,zhangdreamvla,cen2025worldvla}. These limitations hinder robust performance across diverse tasks and out-of-distribution scenarios. Experimental results (Figure~\ref{fig4}) illustrate the difficulties such methods encounter under low-resource settings.

To overcome these challenges and enable seamless integration of generated goal images into existing VLA models, we propose Goal-Injected Fine-Tuning (GIFT), a lightweight and efficient fine-tuning framework. Our approach introduces goal image representations into the observation features through a zero-initialized convolutional layer~\cite{zhang2023adding}, thereby preserving the original input format of the VLA pipeline. The zero-initialized design prevents early interference from goal injection with the pretrained policy, and as training progresses, the parameters of this layer are gradually updated, allowing the goal signal to be progressively incorporated, which enhances goal understanding and improves task performance without compromising model stability. Moreover, we independently train a refined image-editing model for robotic goals, which takes the initial observation and task instruction as input and, through an in-context edit prompt \cite{zhang2025context}, generates a semantically meaningful and visually consistent goal image. This generation occurs prior to action prediction during inference, providing an explicit visual goal image to guide manipulation execution.
Our main contributions are as follows:
\begin{itemize}
\item We propose Goal-Injected Fine-Tuning (GIFT), a lightweight and efficient fine-tuning framework for integrating goal understanding into VLA baselines via a zero convolution without disrupting the manipulation policy.
\item We propose a refined Image Editing for Robotic Goal, incorporating in-context edit prompts to ensure alignment with instructions and a feature injection mechanism to preserve consistency with the observation scene.
\item Extensive experiments demonstrate that GIFT consistently improves VLA performance as shown in Figure~\ref{fig1}, achieving 6.0\% and 13.4\% across two SIMPLER settings, and 4.7\% on LIBERO with only a single fine-tuning epoch.
\end{itemize}

\section{Related Works}

\paragraph{Vision–Language–Action.} 
Vision-Language-Action (VLA) models have emerged as a unified framework integrating visual perception, language understanding, and action execution in robot learning~\cite{team2026motubrain,lin2025hif,shukor2025smolvla,liu2025hybridvla,wang2025bitvla}. Early methods combined pretrained Vision-Language Models (VLMs) with manipulation policies to enable semantic grounding, while recent advances in large-scale multimodal pretraining and some high-quality robot demonstration datasets have substantially improved VLA generalization. Representative approaches, including the RT series~\cite{zitkovich2023rt,belkhale2024rt}, PI series~\cite{black2024pi_0,intelligence2025pi_,intelligence2025pi} and OpenVLA~\cite{kim2025fine,kim2024openvla}, discretize actions and employ autoregressive modeling to predict robot actions, achieving strong performance in familiar environments and under complex instructions. However, their adaptability remains limited by the scarcity and heterogeneity of robotic data, resulting in degraded performance in out-of-distribution settings.

\paragraph{Generative Models for Robotic Control.}
Previous research has investigated integrating generative models into robotic control. Leveraging highly expressive generative models pretrained on internet-scale data, these methods have been applied to low-level control~\cite{zawalskirobotic}, target detection~\cite{peng2024learning}, semantic planning~\cite{huang2025enerverse}, and visual planning~\cite{chen2025goal,hatch2024ghil}. A representative line of work~\cite{bharadhwajgen2act} generates intermediate subgoal images from language instructions using diffusion models and conditions downstream control policies on these subgoals to guide action execution. Despite their promising results, such pipelines are largely incompatible with mainstream VLA architectures, as incorporating goal-conditioned components into existing VLA baselines typically requires full system retraining~\cite{team2026motubrain,zhao2025cot,zhangdreamvla}, incurring substantial computational overhead.

\paragraph{Diffusion-Based Instructional Image Editing.}
Text-to-image diffusion models have demonstrated strong semantic alignment capabilities, enabling the generation of visually consistent images from textual instructions. A range of image editing methods~\cite{yu2025anyedit,gholami2025streamlining} have been proposed, which enable instruction-driven image modification by leveraging pretrained diffusion models. Despite efforts to extend instruction-driven editors to finer-grained datasets and architectures, methods such as Emu Edit~\cite{sheynin2024emu}, OmniGen~\cite{xiao2025omnigen}, and ICEdit~\cite{zhang2025context} continue to suffer from inconsistent generation quality. Recent works~\cite{wu2024turboedit,liu2025step1x,nguyen2025swiftedit} further explore accelerated diffusion models that perform editing within 3–4 denoising steps, substantially improving inference efficiency, but often at the cost of reduced visual fidelity or limited fine-grained control. These limitations are especially evident in the interactive image editing and robotics settings, highlighting the need for image editing methods that are simple, efficient, and robust enough to handle a wide range of robotic scenarios.

\section{Method}
In this section, we present an efficient Goal-Injected Fine-Tuning (GIFT) framework for Vision-Language-Action models, designed to enable rapid integration and low-overhead adaptation of generated goal images into existing baselines. We begin by introducing the problem formulation and the overall framework. Then we detail the proposed fine-tuning framework through which generated goal-image conditions are integrated into the VLA baseline. Finally, we describe our image editing strategy for generating aligned goal images based on the initial observation and instruction. 

\begin{figure}[t]
  \centering
  \includegraphics[width=\columnwidth]{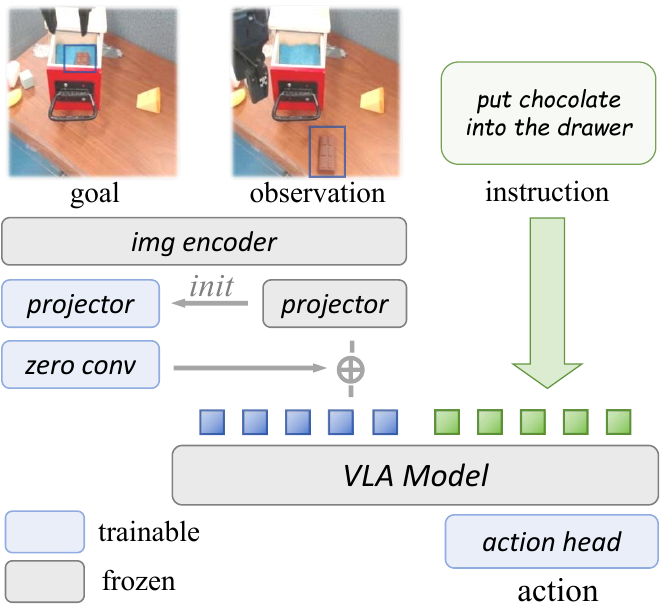}
  \caption{
  Overview of our proposed Goal-Injected Fine-Tuning framework. The goal image is encoded and injected into the VLA model via zero convolution, enabling efficient goal-image condition.
  }
  \label{fig2}
  \vspace{-8pt}
\end{figure}

\subsection{Framework Overview}

We structure the training data for VLA fine-tuning as tuples of the format $D = \{(l, a_{1...T}, o_{1...T})\}$, where $l$ denotes a natural language instruction, $a_{1...T} = \{a_1, ..., a_T\}$ represents the corresponding sequence of robot actions, and $o_{1...T} = \{o_1, ..., o_T\}$ denotes the sequence of visual observations. 
This formulation captures the temporal and multimodal structure of robotic demonstrations, enabling the VLA model to jointly learn perception and action prediction in a unified framework.

\paragraph{VLA.} Vanilla VLA approaches typically fine-tune a pretrained VLM, $\pi_{\phi}$, learning to predict actions $\hat{a}_{t}$ directly based on the current observation $o_t$ and the language instruction $l$ :
\begin{equation}
\hat{a}_t \sim \pi_{\phi}(a_t \mid o_t, l)
\end{equation}

\paragraph{GIFT.} Our core objective is to efficiently incorporate goal image understanding into VLA models. Our method operates in two clearly sequential stages. First, we employ an image editing model $p_\theta$ to generate a visually grounded goal image $\hat{g}$ conditioned on the given initial observation $o_0$ and the task instruction $l$, formulated as:
\begin{equation}
\hat{g} \sim p_\theta(g \mid o_0, l)
\end{equation}
Then, a lightweight goal-injected fine-tuned VLA policy $\pi_{\phi'}$, derived from the pretrained VLA model $\pi_{\phi}$ with only minimal additional parameters, predicts the action $\hat{a}_t$ conditioned on the current observation $o_t$, the language instruction $l$, and the generated goal image $\hat{g}$, formulated as:
\begin{equation}
\hat{a}_t \sim \pi_{\phi'}(a_t \mid o_t, l, \hat{g})
\end{equation}
This two-stage pipeline enables the model to explicitly reason about the desired future states prior to action prediction, thereby strengthening goal alignment and further improving manipulation reliability and consistency across tasks.

\subsection{Goal-Injected Fine-Tuning}
\label{sec:method}

Our proposed method enhances goal-awareness in VLA models through injecting goal image representations, as clearly illustrated in Figure~\ref{fig2}. Instead of training a new manipulation policy from scratch, our approach aims to efficiently adapt pretrained models to better understand and leverage goal images for consistently improved task performance.

At first, we employ a shared pretrained image encoder to extract features from the goal image $\hat{g}$ and the current observation $o_t$, yielding feature embeddings $F_g$ and $F_t$, respectively. Directly injecting the fixed goal feature $F_g$ into the model would act as a constant offset to the observation representation, introducing a static bias that severely limits the model’s representational capacity and adaptability. To address this, we compute the semantic difference between the two embeddings, enhancing goal conditioning, formulated as:
\begin{equation}
F_\Delta = F_g - F_t.
\label{eq}
\end{equation}

Inspired by ControlNet~\cite{zhang2023adding}, we introduce the goal image features into the observation via a zero-initialized convolution which progressively grows the parameters from zero while preventing disruptive signals from affecting the pretrained policy during fine-tuning. Because of its zero initialization, the layer initially suppresses the influence of the goal image, preventing interference with the policy and maintaining stability in early training stages. As training progresses, the layer is gradually updated, allowing goal signals to be integrated into the model in a progressive and controllable manner, which enhances goal comprehension while preserving model stability and training efficiency. The resulting residual feature is integrated into the observation to produce the final embedding $E_{t}$, formulated as:
\begin{equation}
E_{t} =\mathrm{ZeroConv}(\mathrm{Proj_g}(F_\Delta)) + \mathrm{Proj_o}(F_t)
\end{equation}
where $\mathrm{Proj}_g$ and $\mathrm{Proj}_o$ denote the projection layers that map \( F_\Delta \) and \( F_t \) into a shared and aligned feature space, and $\mathrm{ZeroConv}$ represents a zero-initialized convolutional layer.

During fine-tuning, we freeze the vast majority of parameters in the pretrained VLA baseline and update only three lightweight components: the goal-image projector whose parameters are initialized from the observation-image projector, the zero-initialized convolution layer, and the action prediction head. This design enables the model to efficiently adapt to goal-conditioning, improving its capacity about robotic actions.

\subsection{Image Editing for Robotic Goal}
\begin{figure}[t]
  \centering
  \includegraphics[width=\columnwidth]{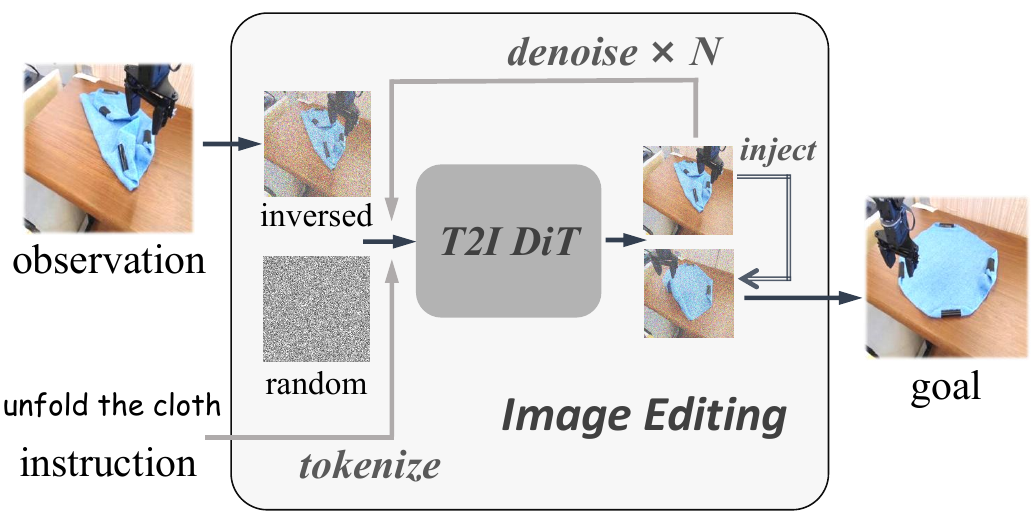}
  \caption{
  Illustration of the proposed Image Editing for Robotic Goal pipeline. 
  Given an observation and instruction, the model combines the inversed latent image with random noise in a T2I DiT and iteratively denoises them to generate an instruction-aligned goal image.
  }
  \label{fig3}
  \vspace{-8pt}
\end{figure}
We propose an image editing approach tailored for robot visual planning tasks. As illustrated in Figure~\ref{fig3}, our method generates a goal image that is semantically aligned with the task instruction, based on the initial observation.

To help the model better understand instructions and improve the quality of edited goal results, we add task instructions to a generative prompt designed for in-context editing~\cite{zhang2025context}. Specifically, for each task instruction, we construct a descriptive prompt in the form:
\begin{quote}
``A side-by-side image of the same scene: the left depicts the observation of our robot, while the right mirrors the left but the observation is obtained after completing the task specified by \{task instruction\}.''
\end{quote}

\begin{table*}[t]
\small
\centering
\renewcommand{\arraystretch}{1.1}
\caption{Comparison of our approach with existing VLA models on the Google robot across four tasks in two SIMPLER settings. 
Both settings include four tasks: (1) Pick coke can, (2) Move near, (3) Open/close drawer, and (4) Open top drawer and place apple. 
}
\setlength{\tabcolsep}{2pt}
\begin{tabular}{p{2.5cm}<{\raggedright}|
p{1.1cm}<{\centering}p{1.1cm}<{\centering}p{1.1cm}<{\centering}p{1.6cm}<{\centering}p{1.05cm}<{\centering}|
p{1.1cm}<{\centering}p{1.1cm}<{\centering}p{1.1cm}<{\centering}p{1.6cm}<{\centering}p{1.05cm}<{\centering}}
\toprule
\multirow{2}{*}{\textbf{Method}} & \multicolumn{5}{c|}{\textbf{Visual Matching}} & \multicolumn{5}{c}{\textbf{Variant Aggregation}} \\
& \textbf{Grasp} & \textbf{Move} & \textbf{Drawer} & \textbf{PutInDrawer}  & \textbf{Avg.} 
& \textbf{Grasp} & \textbf{Move} & \textbf{Drawer} & \textbf{PutInDrawer}  & \textbf{Avg.} \\
\midrule
RT-1          & 85.7 & 44.2  & 73.0 &  6.5 & 52.4 & 89.8  & 50.0 & 32.2 &  2.6 & 43.7  \\
RT-1-X        & 56.7 & 31.7  & 59.7 & 21.3 & 42.4 & 49.0  & 32.3 & 29.4 & 10.1 & 30.7  \\
RT-2-X        & 78.7 & 77.9  & 25.0 &  3.7 & 46.3 & 82.3  & 79.2 & 35.4 & 20.6 & 54.4  \\
Octo-Base     & 17.0 &  4.2  & 22.7 &  0.0 & 11.0 &  0.6  &  3.1 &  1.1 &  0.0 &  1.2 \\
OpenVLA       & 18.0 & 56.3  & 63.0 &  0.0 & 34.3 & 60.8  & 67.7 & 28.8 &  0.0 & 39.3  \\
OpenVLA+GIFT  & 39.7 & 66.0  & 74.9 &  0.0 & 45.1 & 65.9  & 70.7 & 55.2 &  0.0 & 48.0 \\
SpatialVLA    & 81.0 & 69.6  & 57.4 &  1.9 & 52.5 & 84.4  & 71.5 & 34.1 & 11.1 & 50.3  \\
SpatialVLA+GIFT &89.7 &80.6  & 65.9 & 15.3 & 62.9 & 88.9  & 79.7 & 42.8 & 19.5 & 57.5 \\
CogACT        & \textbf{91.3} & 85.0  & 71.8 & 50.9 & 74.5 & \textbf{89.6}  & 80.8 & 28.3 & 46.6 & 61.3 \\
\rowcolor{blue!8} 
CogACT+GIFT   & 91.0 & \textbf{89.9}  & \textbf{78.6} & \textbf{62.6} & \textbf{80.5} & 87.6  & \textbf{86.9} & \textbf{64.7} & \textbf{59.7} & \textbf{74.7} \\
\bottomrule
\end{tabular}

\label{table1}
\vspace{-4pt}
\end{table*}

Our image editing pipeline is clearly illustrated in Figure~\ref{fig3}. We first perform latent inversion \cite{wang2024taming} on the observation image to obtain its corresponding latent representation in latent space. The resulting representation is then concatenated side-by-side with random noise as goal representation, forming a joint latent input that is subsequently fed into the text-to-image diffusion Transformer model (T2I DiT). Through denoising, the model gradually produces a clear and instruction-aligned goal image. To further maintain structural alignment between the generated goal image and the original observation, we introduce an Observation-structure injection mechanism as indicated by the dashed arrow on the right side of Figure~\ref{fig3} during the early denoising stages. The injection is implemented via a weighted fusion of the observation and goal features:
\begin{equation}
\hat {z}_t^{o} = \alpha \cdot z_t^{g} + (1 - \alpha) \cdot z_t^{o}
\end{equation}
where \( z_t^{g} \) denotes the latent feature of the goal at timestep \( t \), \( z_t^{o} \) is the corresponding observation feature, and \( \alpha \in [0, 1] \) is a blending coefficient.

\section{Experiments}
To evaluate the effectiveness of our approach, we conduct experiments on two simulation benchmarks and real-world robotic tasks. Section~\ref{4.1} introduces the evaluation environments and implementation details. Section~\ref{4.2} presents performance comparisons with existing VLA models, along with qualitative analyses of goal image editing and fine-tuning efficiency. Section~\ref{4.3} reports ablation studies to isolate the contributions of individual framework components.
\subsection{Experimental Setup}
\label{4.1}
We conduct evaluations on both simulation and real-world robotic tasks. Each task is tested across diverse initial states, and overall success rates (SR) are reported to systematically assess our method.

\paragraph{Evaluation Benchmarks.}
We evaluate our method on two widely used robotic simulation benchmarks. SIMPLER~\cite{li2025evaluating} includes Visual Matching, aligning simulation visuals with real scenes, and Variant Aggregations, introducing variations in background, lighting, distractors, and textures. LIBERO~\cite{liu2023libero} targets the Franka Emika Panda robot with four task suites—Spatial, Object, Goal, and Long. 

\paragraph{Real-World.} 
We employ a 6-DOF Synria Alicia-D equipped with a 1-DOF gripper. Both the wrist-view and third-view images are captured using Intel RealSense D405 cameras as illustrated in Figure~\ref{fig:twopdf}. The robotic system is tested on tasks such as ``pick up \textless obj\textgreater'' and ``put \textless obj\textgreater  into basket''. Out-of-distribution (OOD) refers to scenarios with different lighting conditions, used to evaluate the generalization of our approach. For each task under each evaluation setting, we conduct 30 trials and report the task success rate.

\paragraph{Training Details.} For simulation experiments, we incorporate the proposed GIFT into three baselines, OpenVLA~\cite{kim2024openvla}, CogACT~\cite{li2024cogact} and SpatialVLA~\cite{qu2025spatialvla}. To accurately evaluate adaptation performance, we strictly align the fine-tuning datasets with their corresponding robotic platforms: models evaluated on the WidowX platform in SIMPLER are fine-tuned on the BridgeDataset V2, while those evaluated on the Google Robot platform in SIMPLER are fine-tuned on the Fractal dataset. For evaluations on the LIBERO benchmark, models are specifically fine-tuned on the LIBERO dataset. All models are fine-tuned for a single epoch. Subsequently, we fine-tune the image editing DiT model on a processed version of these three dataset. For real-world experiments, we adopt $\pi_{0.5}$~\cite{intelligence2025pi_} as the baseline. We sample data for two tasks with 30 demos each using the LeRobot format. We perform LoRA fine-tuning on $\pi_{0.5}$ as well as the GIFT variant. See more details in the Appendix~\ref{TrainingDetails}.
\begin{table}[t]
\centering
\small
\caption{Evaluation results on the WidowX robot in the SIMPLER \textit{Visual Matching} setting. Four tasks are: (1) Put spoon on towel (2) Put carrot on plate (3) Stack green block on yellow block and (4) Put eggplant in yellow basket. Each task is evaluated 3 times.}
\begin{tabular}{p{2.3cm}<{\raggedright}|p{0.5cm}<{\centering}p{0.5cm}<{\centering}p{0.6cm}<{\centering}p{0.7cm}<{\centering}|p{0.5cm}<{\centering}} 
\toprule
\textbf{Method} & \textbf{Task1} & \textbf{Task2} & \textbf{Task3} & \textbf{Task4} & \textbf{Avg.} \\
\midrule
 RT-1-X                    & 0.0  & 4.2  & 0.0  & 0.0  & 1.1 \\
 Octo-Base                 & 15.8 & 12.5 & 0.0  & 41.7 & 17.5 \\
SuSIE	                   & 34.7 & 45.5 & 6.9	& 40.2	& 31.9 \\
GHIL-Glue	               & 54.2 &	48.6 & 11.1	& 47.2	& 40.3 \\
 OpenVLA                   & 4.2  & 0.0  & 0.0  & 12.5 & 4.2 \\
 OpenVLA+GIFT              & 9.7  & 0.0  & 0.0  & 31.7 & 10.4 \\
 SpatialVLA                & 16.7 & 25.0 & 25.0 & 66.7 & 33.3 \\
 SpatialVLA+GIFT           & 34.7 & 37.5 & 29.2 & 72.2 & 43.4 \\
 CogACT                    & 71.7 & 50.8 & 15.0 & 67.5 & 51.3 \\
\rowcolor{blue!8} 
 CogACT+GIFT               & \textbf{80.6} & \textbf{63.9} & \textbf{43.1} &\textbf{83.3}  & \textbf{67.7} \\
\bottomrule
\end{tabular}
\label{table2}
\vspace{-8pt}
\end{table}
\subsection{Experimental Results}
\label{4.2}
\paragraph{SIMPLER.} 

We present a quantitative comparison between our method and existing VLA models in Table~\ref{table1} across four tasks under two SIMPLER evaluation settings: Visual Matching (VM) and Variant Aggregation (VA). Incorporating GIFT into OpenVLA, SpatialVLA, and CogACT consistently improves performance across both settings. In VM, CogACT+GIFT achieves the best average success rate of 80.5\%, improving over CogACT by 6.0\%. In the more challenging VA setting, CogACT+GIFT reaches 74.7\%, outperforming CogACT by 13.4\%. We further evaluate CogACT+GIFT on the WidowX robot under the SIMPLER VM setting, as shown in Table~\ref{table2}. GIFT improves CogACT across all four tasks, increasing the average success rate from 51.3\% to 67.7\%. The gain is especially large on Task 3, where performance improves from 16.4\% to 43.1\%.

\paragraph{LIBERO.}
We present quantitative comparisons with VLA models on four tasks (Spatial, Object, Goal, and Long) under the LIBERO benchmark in Table~\ref{table3}. Overall, OpenVLA+GIFT achieves the best performance across all tasks, with an average success rate of 81.2\%, improving upon OpenVLA by 4.7\% and demonstrating GIFT’s effectiveness in multi-task settings. Notably, OpenVLA+GIFT attains strong results on the Spatial and Goal tasks, reaching 89.5\% and 88.6\%, respectively, which is significantly better than other methods. On the challenging LIBERO-Long task, it also achieves the highest success rate of 56.0\%, indicating its robustness in complex long-term planning tasks.

\begin{figure}[t]
    \centering
    \includegraphics[height=0.12\textheight]{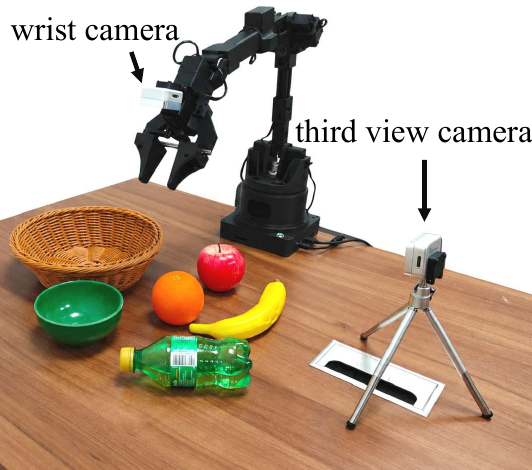}%
    \raisebox{0mm}{\includegraphics[height=0.13\textheight]{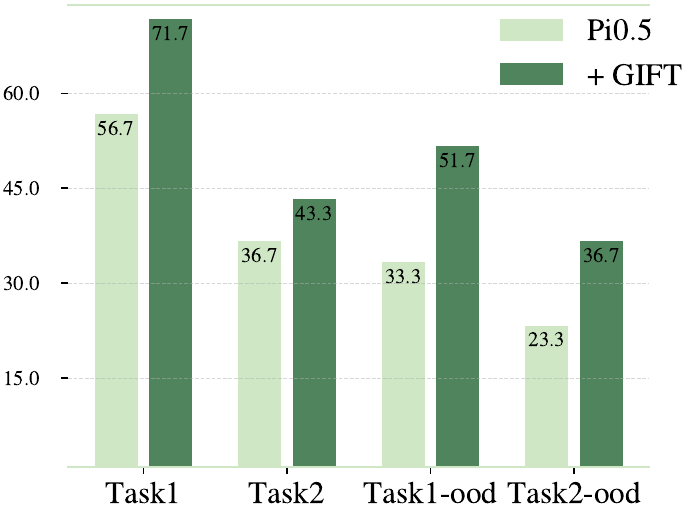}}
    \caption{Real-world deployment setup and evaluation results. Left: Real-world experimental environment. Right: Task performance comparison between $\pi_{0.5}$ and the variant with GIFT across ID and OOD scenarios.}
    \label{fig:twopdf}
    \vspace{-8pt}
\end{figure}

\paragraph{Real-World.} 
To evaluate whether our method remains effective in real-world deployment, we conduct real-robot experiments under both in-domain and out-of-distribution (OOD) settings, where the OOD setting specifically refers to dim-lighting conditions. The results in Figure~\ref{fig:twopdf} show the performance comparison between the baseline $\pi_{0.5}$~\cite{intelligence2025pi_} and the variant enhanced with GIFT across in-domain (ID) and out-of-distribution (OOD) settings. For in-domain tasks, GIFT improves the success rate from 56.7\% to 71.7\% on Task1 and from 36.7\% to 43.3\% on Task2. For OOD tasks with varying lighting conditions, the gains are more pronounced: Task1-ood improves from 33.3\% to 51.7\%, and Task2-ood improves from 23.3\% to 36.7\%. These results demonstrate that GIFT enhances task performance and robustness under both familiar and novel conditions.

\paragraph{VLM Backbone Generalization.} 
To evaluate the generalization of our proposed fine-tuning strategy across different VLM backbones, we apply GIFT to OpenVLA and CogACT, which are built upon the Prismatic VLM~\cite{karamcheti2024prismatic}, as well as to SpatialVLA, which is based on PaLI-Gemma2~\cite{steiner2024paligemma}. Despite relying on fundamentally different backbone architectures, GIFT consistently improves performance across both settings significantly, increasing the average success rate of CogACT by 6.0\% and 13.4\% under the Visual Matching and Variant Aggregation settings, respectively, and yielding corresponding gains of 10.4\% and 7.2\% for SpatialVLA, thereby demonstrating strong model-agnostic effectiveness and robustness in practice.

\begin{table}[t]
\caption{Evaluation results on the Franka Emika Panda robot in the LIBERO benchmark across four task suites: Spatial, Object, Goal, and Long-horizon tasks.} 
\centering
\footnotesize
\begin{tabular}{p{2.3cm}<{\raggedright}|p{0.7cm}<{\centering}p{0.7cm}<{\centering}p{0.4cm}<{\centering}p{0.6cm}<{\centering}|p{0.4cm}<{\centering}} 
\toprule
\textbf{Method}  & \textbf{Spatial} & \textbf{Object} & \textbf{Goal} & \textbf{Long} & \textbf{Avg.}\\
\midrule
Diffusion Policy   & 78.3  & \textbf{92.5} & 68.3 & 50.5 & 72.4\\
Octo-Base          & 78.9  & 85.7  & 84.6  & 51.1 & 75.1 \\
TraceVLA           & 84.6  & 85.2  & 75.1  & 54.1 & 74.8  \\
OpenVLA            & 84.7  & 88.4  & 79.2  & 53.7 & 76.5  \\
\rowcolor{blue!8} 
OpenVLA+GIFT            & \textbf{89.5} & 90.6 & \textbf{88.6} & \textbf{56.0} & \textbf{81.2}\\
\bottomrule
\end{tabular}
\label{table3}
\end{table}



\begin{figure*}[t]
  \centering
  \includegraphics[width=\textwidth]{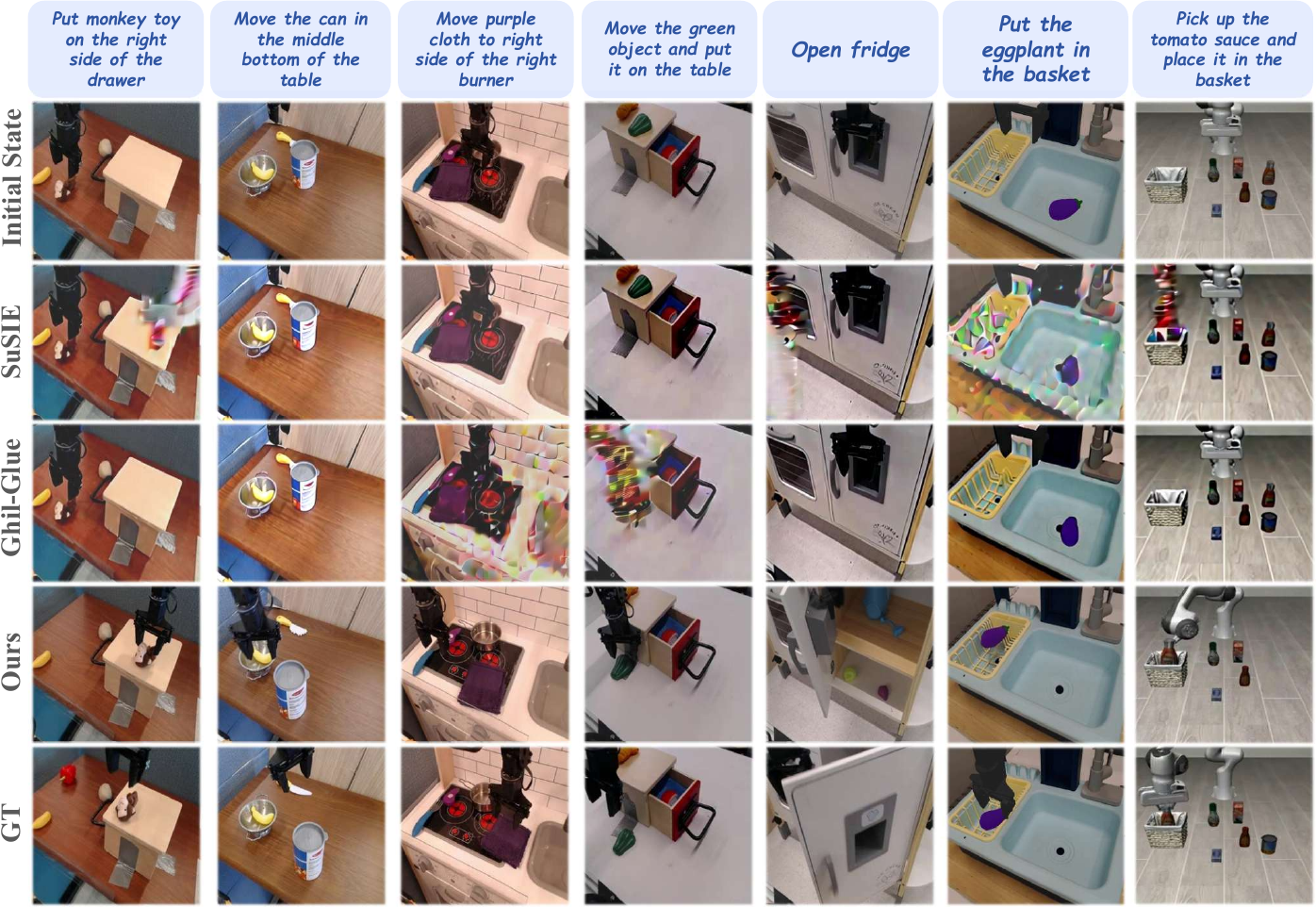}
  \caption{Qualitative comparison of visual editing results across various goal-directed manipulation tasks, comparing our method with existing approaches (SuSIE and Ghil-Glue). From top to bottom, the rows show: the instruction, the initial state, the edited result generated by SuSIE, Ghil-Glue, our method, and the ground-truth (GT) image. }
  \label{fig5}
  \vspace{-8pt}
\end{figure*}

\paragraph{GIFT Enhances Rather Than Creates Capabilities.}
Although GIFT consistently improves performance across benchmarks, its gains are best understood as strengthening capabilities already partially present in the pretrained policy, rather than introducing entirely new manipulation skills. For example, OpenVLA remains at 0\% success on PutInDrawer after goal injection, despite achieving clear average improvements in SIMPLER. This suggests that explicit visual goal conditioning alone cannot compensate for a complete lack of underlying task competence in the base model. Instead, by preserving most pretrained parameters while injecting goal-aware guidance, GIFT serves as an efficient capability amplifier that improves goal understanding and execution alignment within the behavioral scope of the original VLA policy.

\paragraph{Image Editing Results.}
We present qualitative comparisons of visual editing results in Figure~\ref{fig5} against two existing methods, SuSIE~\cite{blackzero} and Ghil-Glue~\cite{hatch2024ghil}, across multiple tasks. Our method consistently produces goal images that better align with task instructions, exhibiting improved object localization, semantic coherence, and structural integrity. In contrast, SuSIE and Ghil-Glue often fail to generate instruction-aligned edits, introducing artifacts such as object misplacement, unrealistic textures, and structural distortions. For example, in the task \textit{``Open fridge''}, both SuSIE and Ghil-Glue are unable to accurately depict the fridge door as open, while our method successfully renders the intended transformation correctly, closely matching the expected outcome. The results demonstrate that our image editing framework achieves superior alignment with task instructions and produces realistic outputs, particularly under robotic conditions.

\begin{figure}[t]
  \centering
  \includegraphics[width=\columnwidth]{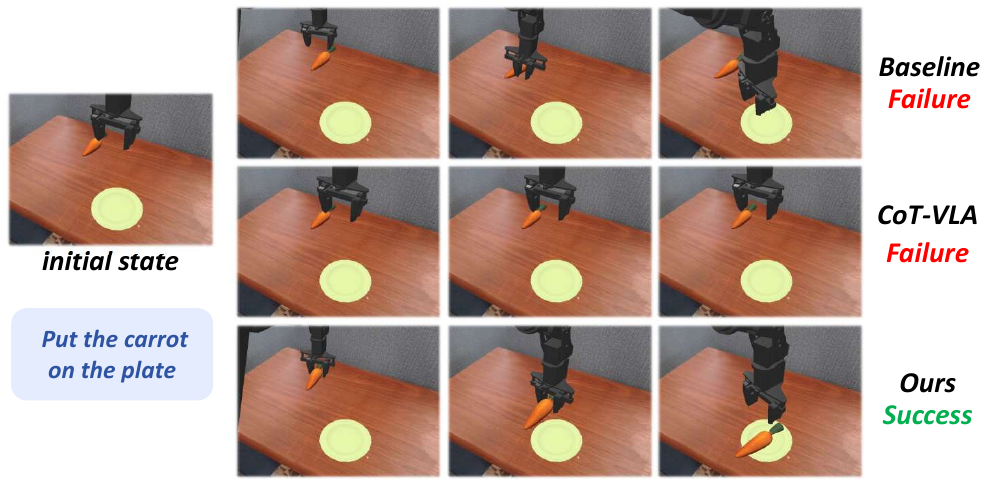}
  \caption{
  Qualitative comparison among the baseline, GIFT (Ours), and CoT-VLA on the task “put the carrot on the plate,” where both GIFT and CoT-VLA are trained on the same baseline for only a single epoch.
  }
  \label{fig4}
\end{figure}

\paragraph{Fine-Tuning Efficiency.}
To evaluate the efficiency of our fine-tuning and its ability to preserve pretrained policies, we reproduce the CoT-VLA~\cite{zhao2025cot} framework on the same dataset for a single epoch(48 hours). Figure~\ref{fig4} shows a qualitative comparison on the task \textit{``put the carrot on the plate''} among our method, CoT-VLA, and the baseline (CogACT). The baseline attempts the task but fails, CoT-VLA shows almost no meaningful behavior, while our method successfully completes it. These results highlight the effectiveness of our approach in low-resource settings, where preserving pretrained priors and incorporating goal guidance improve performance.

\subsection{Ablation Study}
\label{4.3}
\begin{figure}[t]
  \centering
  \includegraphics[width=\columnwidth]{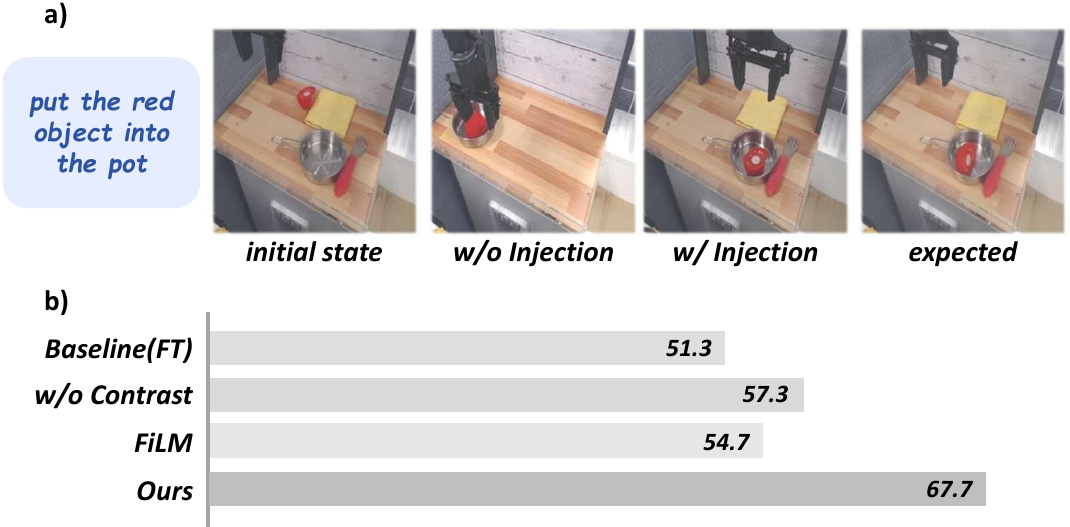}
  \caption{
  Ablation study results. (a) shows the effect of Observation-structure Injection on visual editing. (b) presents quantitative comparisons on SIMPLER Visual Matching across four configurations: baseline, GIFT without Semantic Contrast, GIFT with ZeroConv replaced by FiLM, and GIFT.
  }
  \label{fig6}
  \vspace{-8pt}
\end{figure}

\paragraph{Observation-structure Injection.}
We present a qualitative comparison in Figure~\ref{fig6}(a) to illustrate the effect of the Observation-Structure Injection mechanism on visual editing performance for the task “Put the red object into the pot.” By injecting the goal image representation into the observation representation, the model more reliably and precisely infers the intended end state while preserving structural consistency in the scene layout between the two images across complex scenes. In contrast, although the variant without injection is still able to complete the task, it often exhibits inconsistencies, such as diminished retention of secondary objects and variations in scene brightness.

\paragraph{Semantic Contrast.}
We present a quantitative comparison in Figure~\ref{fig6}(b) across different configurations under the SIMPLER \textit{Visual Matching} setting. Applying GIFT without Semantic Contrast (Eq.~\ref{eq}) yields a modest performance improvement, primarily by introducing a stable visual prior. In contrast, the Semantic Contrast mechanism explicitly computes feature-level differences between the goal and observation images at the representation space, allowing the model to more effectively capture task-relevant semantic changes. This enhancement substantially improves the model’s goal understanding capability, resulting in an overall performance increase from 57.5\% to 67.7\%.

\begin{table}[t]
\centering
\caption{Comparison of GIFT success rates (\%) under varying goal image qualities. Evaluated tasks span across the SIMPLER (first five rows) and LIBERO (last four rows) benchmarks. The GIFT injection module is fixed across all settings.}
\label{tab:goal_quality}
\resizebox{\columnwidth}{!}{%
\begin{tabular}{l|ccccc}
\toprule
\textbf{Task} & \textbf{Degraded} & \textbf{SuSIE} & \textbf{Turbo} & \textbf{Ours} & \textbf{GT Goal} \\
\midrule
Grasp      & 52.1 & 66.7 & 80.0 & 87.6 & 90.3 \\
Move   & 45.7 & 55.9 & 79.6 & 88.4 & 88.9 \\
Drawer        & 25.5 & 29.6 & 70.9 & 71.7 & 75.3 \\
Place     & 41.1 & 35.6 & 72.1 & 75.9 & 80.5 \\
Stack      & 10.6 & 0.0  & 38.3 & 41.1 & 52.5 \\
\midrule
Spatial       & 50.4 & 48.6 & 85.6 & 89.5 & 89.9 \\
Object        & 58.4 & 60.5 & 85.9 & 90.6 & 92.1 \\
Goal          & 55.1 & 58.1 & 84.1 & 88.6 & 88.1 \\
Long          & 24.0 & 20.9 & 55.7 & 56.0 & 56.5 \\
\bottomrule
\end{tabular}%
}
\end{table}
\paragraph{Goal Injection Mechanisms.}
We perform an ablation study on the goal injection module by replacing the proposed $\mathrm{ZeroConv}$ with alternative designs, including cross-attention and FiLM, as shown in Figure~\ref{fig6}(b). Introducing cross-attention directly disrupts the pretrained policy, leading to severe optimization instability and near-zero success rates. FiLM preserves the original architecture but underperforms ZeroConv, as its channel-wise scaling induces feature drift in multimodal embeddings, weakening goal-conditioned representations and reducing fine-tuning stability and efficiency.

\paragraph{Better Editing Helps.}
As shown in Table \ref{tab:goal_quality}, we present a comparison of GIFT under varying goal image qualities across diverse task families in both the SIMPLER and LIBERO benchmarks, utilizing CogACT and OpenVLA as their policies. By fixing the injection module and varying the goal source, we find that degraded goals severely hurt performance. Off-the-shelf editors like SuSIE lack spatial coherence (e.g., yielding 0.0\% on Stacking, 20.9\% on LIBERO-Long), and the Turbo Version sacrifices visual fidelity for speed ($\sim$1.1s). Conversely, our method ($\sim$4.2s one-time cost) outperforms all generated baselines across all nine tasks, nearly matching the ground-truth (GT) upper bound. This demonstrates that high-quality goal generation is crucial for reliable execution, and our approach optimally balances efficiency and robustness.


\section{Conclusion}
In this work, we propose Goal-Injected Fine-Tuning (GIFT), a lightweight framework for efficiently integrating generated goal images into pretrained VLA models. GIFT injects goal features via a zero-initialized convolution, ensuring stable adaptation without disrupting the pretrained policy. Additionally, we introduce a refined image editing method to generate semantically and visually aligned goal images from observations and task instructions. Experiments demonstrate that our method efficiently enhances goal understanding and policy robustness with minimal fine-tuning.

\section*{Limitations}
Despite its strong empirical performance, our approach has several limitations. First, incorporating goal image editing introduces additional computational overhead and, together with the two-stage separation between goal generation and action prediction, may lead to efficiency bottlenecks and potential distributional mismatches; future work could address this by leveraging faster image editing techniques or unified end-to-end architectures that jointly model goal reasoning and policy learning. Second, since GIFT freezes most parameters of the pretrained VLA model, it primarily enhances goal awareness but cannot extend the model’s capability beyond tasks already supported by the baseline, motivating future exploration of efficient fine-tuning strategies to expand task coverage.

\section*{Acknowledgment}
This research is supported by the National Natural Science Foundation of China (No.62476127,62306140),  the Natural Science Foundation of Jiangsu Province (No.BK20242039), the Fundamental Research Funds for the Central Universities(NO.NXD2026006), Disciplinary Basic Research Project Funding, (No.ILF26023), the Research Fund (No.PO250624101698), the Scientific Research Starting Foundation of Nanjing University of Aeronautics and Astronautics (No.YQR21022), and the High Performance Computing Platform of Nanjing University of Aeronautics and Astronautics.

\bibliography{custom}

\appendix

\clearpage



\section{Training Details}
\label{TrainingDetails}
\subsection{Training Setup and Fine-tuning Details.}

We conduct GIFT fine-tuning under two experimental settings. Unless otherwise specified, all VLA models are fine-tuned for a single epoch with a batch size of 256 and a constant learning rate of $1\mathrm{e}{-5}$. Training is implemented with PyTorch Fully Sharded Data Parallel (FSDP) on 4 NVIDIA L20 GPUs.
For three baseline models evaluated in the SIMPLER environment, the training setups are explicitly decoupled based on the robot embodiment: evaluations under the WidowX setting are trained exclusively on the BridgeDataset V2, whereas evaluations under the Google Robot setting are trained on the Fractal dataset. We fine-tune OpenVLA on the LIBERO dataset specifically for evaluations on the LIBERO benchmark.
The training time is presented in Table~\ref{tab:training_time}.

\begin{table}[h]
\centering
\caption{Training time (hours) for a single epoch across different datasets with 4 NVIDIA L20 GPUs.}
\label{tab:training_time}
\resizebox{\columnwidth}{!}{%
\begin{tabular}{lccc}
\toprule
\textbf{Model} & \textbf{BridgeDataset V2} & \textbf{Fractal} & \textbf{LIBERO} \\
\midrule
CogACT & 24.0 & 43.2 & -- \\
OpenVLA  & 24.0 & 44.5 & 9.6 \\
SpatialVLA  & 20.0 & 36.0 & -- \\
\bottomrule
\end{tabular}%
}
\end{table}


\begin{figure*}[t]
  \centering
  \includegraphics[width=\textwidth]{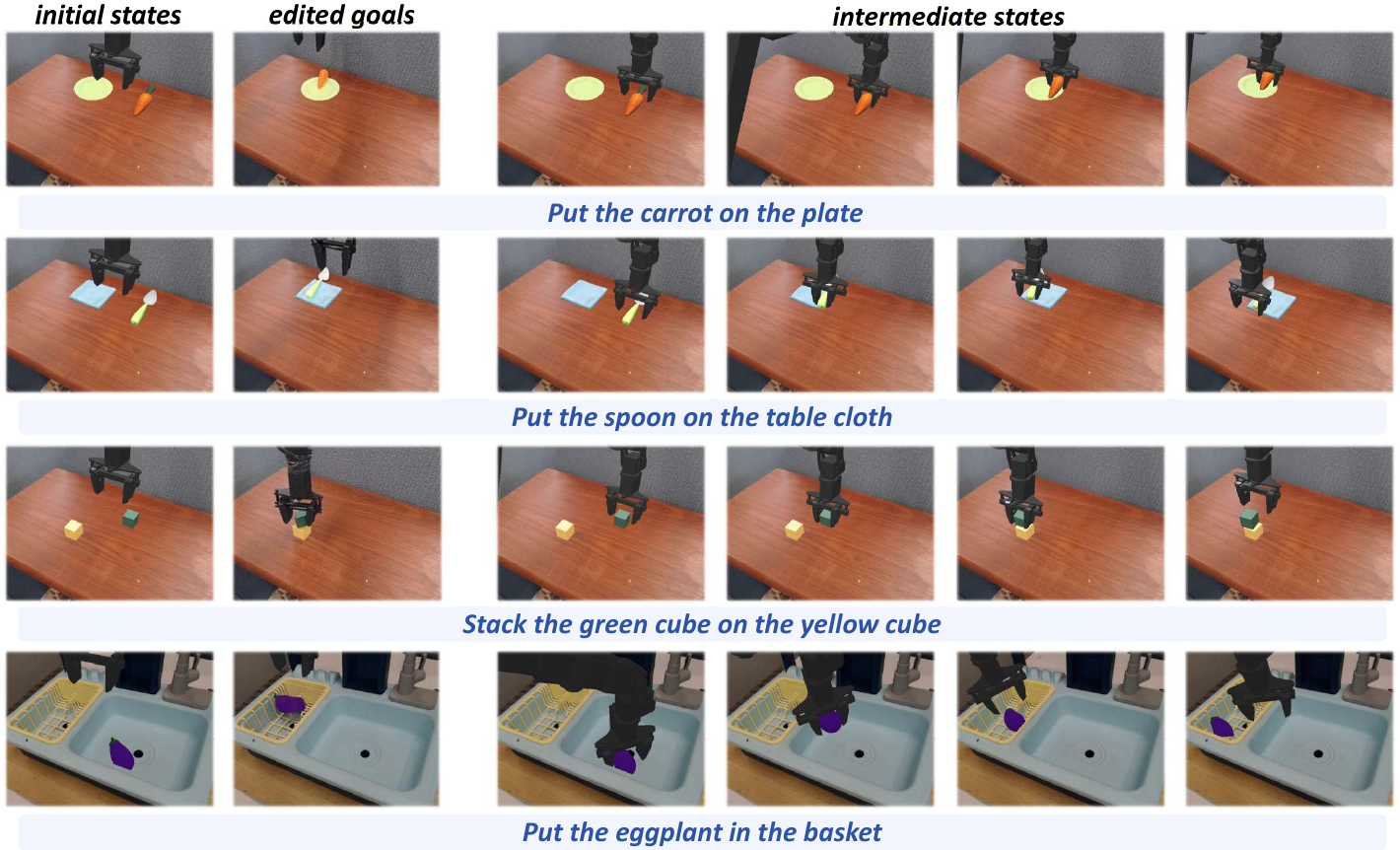}   
\caption{Qualitative results on the SIMPLER simulation with the WidowX robot. Each row illustrates a manipulation task, showing the initial state, the goal image, and intermediate execution frames. Our method generates semantically accurate and visually coherent goal images that align with the task instruction, effectively guiding the robot through various object-centric tasks such as precise placement, stacking, and goal-oriented repositioning.}

  \label{fig10}
  \vspace{-6pt}
\end{figure*}

\subsection{Real-Robot Adaptation and Training Details} 
For real-robot adaptation, we collected demonstrations via teleoperation for two manipulation tasks: ``pick up \textless obj\textgreater'' and ``put \textless obj\textgreater into basket''. For each task, we collected 30 trajectories at 30 Hz, resulting in a compact task-specific dataset with joint-angle action annotations. During data collection, the positions of both the target object and the basket were varied across trajectories. This data collection protocol provides synchronized visual observations, language task prompts, and low-level joint-angle supervision, enabling direct adaptation of the policy to the target robot embodiment.

We first fine-tuned $\pi_{0.5}$ using a joint-angle LoRA fine-tuning strategy. The base model parameters were frozen, and only the LoRA adapters were optimized. Specifically, we used the $\pi_{0.5}$ configuration with an action horizon of 10, continuous state inputs and a LoRA-enabled 300M action expert. The model was initialized from the official $\pi_{0.5}$ base checkpoint, and prompts were derived from task descriptions. We trained the policy for 30,000 steps with a batch size of 20 using AdamW, gradient clipping of 1.0, and a cosine learning-rate schedule with 1,000 warmup steps, a peak learning rate of $5\times10^{-5}$, and a decay horizon of 1,000,000 steps. EMA was disabled to avoid maintaining an additional copy of the parameters during LoRA fine-tuning. This stage was conducted on a single NVIDIA L20 GPU and required approximately 40 hours.
Then, we deployed GIFT on top of the fine-tuned policy checkpoint and further trained the GIFT variant for one epoch. GIFT was injected only into the third-view image branch, where the global scene layout and goal-relevant spatial relations are most visible. The wrist-view branch was left unchanged to preserve local gripper-object interaction cues learned during policy fine-tuning. This additional adaptation stage was lightweight compared with policy fine-tuning, requiring approximately 2 hours on a single NVIDIA L20 GPU. Since GIFT is attached to the execution pipeline as an additional goal-conditioned module, this stage adapts the goal-image conditioning mechanism while preserving the previously learned policy capability.
In parallel, we lightly adapted our image editing model on the third-view image in collected training set for 3 epochs.

\subsection{Image Editing Fine-tuning Details.}
\label{appa2}
To improve robustness across both the SIMPLER and LIBERO benchmarks, we construct a mixed training set using 90\% of BridgeDataset V2, 90\% of Fractal dataset and 30\% of LIBERO, while reserving the remaining 10\% of BridgeDataset V2 for evaluation. We use a blending coefficient $\alpha$ of 0.5, and the denoising process is performed in 20 steps. Observation-structure Injection is applied during the first 25\% of the steps. Each sample consists of the first trajectory frame as the initial observation, the final frame as the goal image, and the corresponding task instruction as the text prompt.
We initialize the image editor from FLUX.1-Fill-dev and fine-tune it in bf16 precision with LoRA. We set the rank to $r=32$, the scaling factor to $\alpha=32$, and use Gaussian initialization for LoRA weights. LoRA modules are applied to the input embedding layer, attention projections, normalization-related linear layers, and feed-forward/output projections in both multi-stream and single-stream transformer blocks, including Q/K/V projections, attention output projections, normalization linear layers, the multi-stream feed-forward output layer, and the single-stream MLP/output projections.
During fine-tuning, both text encoders and the VAE remain frozen, while the DiT backbone is adapted only through the inserted LoRA parameters. The editor is trained for 3 epochs on 4 NVIDIA L20 GPUs with a per-device batch size of 6 and gradient accumulation over 6 steps, resulting in an effective batch size of 144. We enable gradient checkpointing to reduce memory usage. To improve conditional robustness, we apply text-condition dropout and image-condition dropout, each with 0.1.

\begin{figure*}[t]
  \centering
  \includegraphics[width=\textwidth]{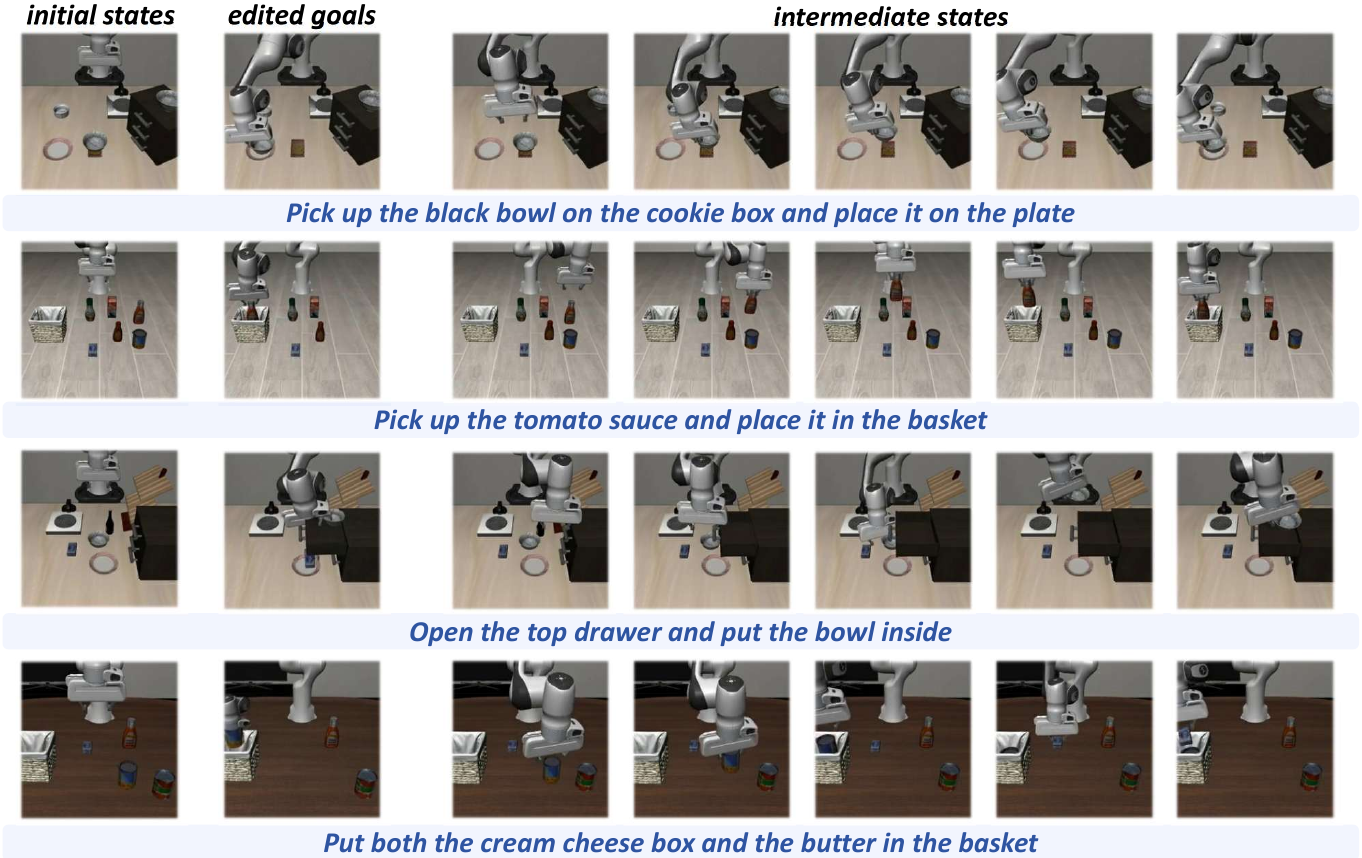}   
\caption{Visualizations of our model's performance across four representative tasks in the LIBERO simulation environment. Each row corresponds to a specific task instruction from one of the four LIBERO suites (Spatial, Object, Goal, and Long). For each task, we show the initial observation, the edited goal image and several intermediate execution frames. Our proposed model demonstrates the ability to produce semantically meaningful goals and successfully guide the robot through multi-step manipulation sequences.}
  \label{fig9}
  \vspace{-10pt}
\end{figure*}
\section{Evaluation Details}

\subsection{SIMPLER}
Evaluating robotic manipulation policies is challenging due to the diversity of hardware platforms, environmental settings, and task definitions across different studies. Although standardized real-world setups can improve comparability, they typically require considerable time and resources. To balance realism and reproducibility, we adopt the SIMPLER benchmark, which enables the evaluation of policies trained on real robot data within high-fidelity simulation environments. This setup allows for fair and efficient benchmarking while preserving alignment with real-world performance. 

We utilize all task variants provided in SIMPLER across two robotic platforms: \textbf{Google} and \textbf{WidowX}. The Google setup includes the following tasks: 
(1) \textit{pick Coke can}, 
(2) \textit{move \{obj1\} near \{obj2\}}, 
(3) \textit{(open / close) (top / middle / bottom) drawer}, and 
(4) \textit{open top drawer; place apple into top drawer}. 
The WidowX setup includes: 
(1) \textit{put the spoon on the towel}, 
(2) \textit{put carrot on plate}, 
(3) \textit{stack the green block on the yellow block}, and 
(4) \textit{put eggplant into yellow basket}. 

For the Google robot setup, both Visual Matching (VM) and Variant Aggregation (VA) evaluations are performed, while only the VM protocol is applied to the WidowX setup. Due to the limited number of original trials on WidowX (24 per task), we repeat each trial five times. For further implementation and evaluation details, we refer readers to the SIMPLER paper \cite{li2025evaluating}.

\subsection{LIBERO}  
The \textbf{LIBERO} benchmark (Lifelong Robot Manipulation Benchmark) \cite{liu2023libero} is designed for studying knowledge transfer in multistep, multitask, and longitudinal robot-learning problems, with 40 procedurally generated language-conditioned tasks grouped into four distinct suites: \emph{LIBERO‑Spatial, LIBERO‑Object, LIBERO‑Goal, LIBERO‑Long}. Each suite contains 10 task instructions, each designed to evaluate a specific type of distribution shift, and for each task instruction, we conduct 50 rollouts to ensure robust statistical evaluation.

All tasks are executed in the high-fidelity Robosuite/MuJoCo simulation environments using PDDL-derived scene layouts and human-teleoperated demonstrations, with each task specified by a natural-language instruction generated from templated goal predicates (for example: "Pick the red block and place it into the green bowl"). At each timestep, agents receive RGB‑D images and proprioceptive readings (joint positions, gripper state) at a 20Hz control frequency consistently.

\subsection{Real World.} 
To evaluate whether our method remains effective in real-world deployment, we conduct real-robot experiments under both in-domain and out-of-distribution (OOD) settings. The system uses a 6-DOF robotic arm equipped with a 1-DOF gripper, with wrist-view and third-view RGB observations captured by Intel RealSense D405 cameras. We evaluate two methods on two manipulation tasks: ``pick up \textless obj\textgreater'' and ``put \textless obj\textgreater into basket''. The target objects include common tabletop items such as bottles, bananas, and apples, as illustrated in Figure~\ref{fig:twopdf}.
For the in-domain setting, experiments are conducted under the same lighting condition as the training data. To assess OOD robustness, we repeat the same tasks under dim lighting while keeping the robot platform, camera setup, workspace layout, object set, and task definitions unchanged. A pick-up trial is considered successful if the robot grasps and stably lifts the specified object. A basket trial is considered successful if the robot picks up the target object and places it inside the basket. For each method, setting, and task, we conduct 30 trials and report the success rate as the percentage of successful executions.


\begin{figure*}[t]
  \centering
  \includegraphics[width=\textwidth]{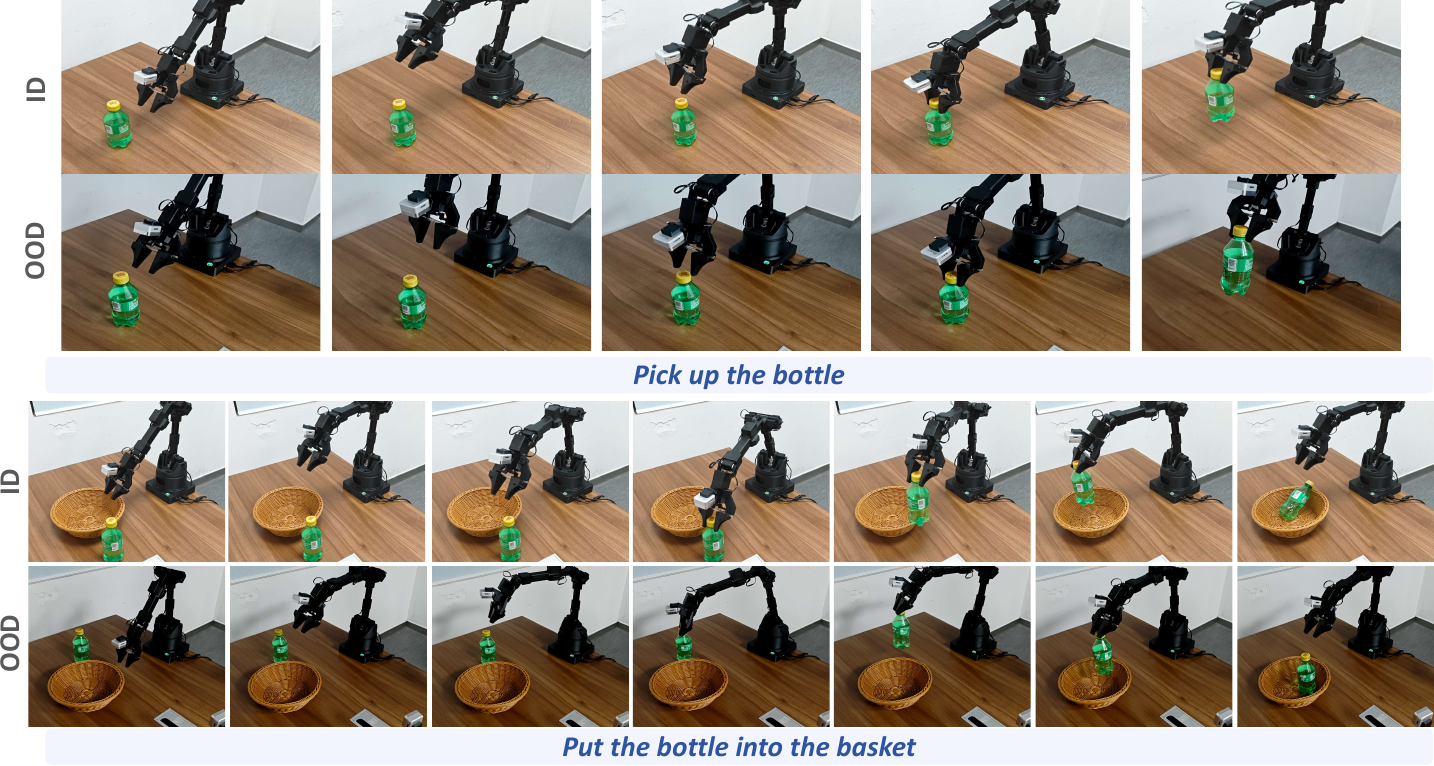}   
\caption{
Qualitative real-robot rollouts under in-domain (ID) and out-of-distribution (OOD) lighting conditions. We show two tasks, ``pick up the bottle'' and ``put the bottle into the basket''. Each row visualizes a sequence of third-view observations during policy execution. The robot successfully completes both tasks under ID lighting and maintains robust goal-directed behavior under dim OOD lighting.
}
\label{fig:real_robot_rollouts}
  \vspace{-10pt}
\end{figure*}

\subsection{Comparison Algorithms}
To comprehensively evaluate the performance of our proposed approach, we conduct comparative evaluations against a diverse set of representative baseline methods, which are detailed below:
\begin{itemize}
    \item \textbf{RT-1} \cite{brohan2022rt}: A Transformer-based robotic policy model trained on large-scale real-world trajectories, which directly maps sequences of images and natural language instructions to tokenized robot actions for manipulation tasks.

    \item \textbf{RT-1-X} \cite{o2024open}: An improved variant of RT-1 trained on the Open X-Embodiment dataset, which enhances cross-robot generalization and robustness.
        
    \item \textbf{RT-2-X} \cite{o2024open}: A vision-language-action model co-trained on web-scale vision-language data and multi-robot trajectories, achieving strong performance and emergent reasoning ability across unseen tasks and robot embodiments.

    \item \textbf{Octo} \cite{ghosh2024octo}: A generalist model trained on the Open-X-Embodiment dataset without using vision-language model initialization. Similar to OpenVLA, we evaluate it using its released checkpoint and apply fine-tuning for downstream tasks.
    
    \item \textbf{Diffusion Policy} \cite{chi2023diffusion}: A state-of-the-art imitation learning algorithm trained from scratch for each test scenario in LIBERO and Franka-Tabletop. It incorporates action chunking and proprioception, and conditions on DistilBERT-based language embeddings.

    \item \textbf{OpenVLA} \cite{kim2024openvla}: An open-source Vision-Language-Action model that fine-tunes pretrained vision-language models on the OpenX dataset. We use its official checkpoint released for Bridge-V2 and fine-tune the model for our LIBERO and Franka-Tabletop experiment settings.

    \item \textbf{CogACT} \cite{li2024cogact}: A foundation VLA model that combines powerful pretrained vision-language-model backbones (e.g. DINOv2, SigLIP, Prismatic) with a dedicated diffusion action transformer.

    \item \textbf{SpatialVLA}\cite{qu2025spatialvla} : A spatially enhanced VLA model that incorporates egocentric 3D position encoding and adaptive action grids to improve spatial understanding and generalist manipulation policy learning across diverse robotic tasks and embodiments.

    \item \textbf{TraceVLA}\cite{zhengtracevla} : A generalist VLA model that leverages trajectory-level action traces to enhance long-horizon reasoning and temporal consistency, enabling more robust policy learning across diverse robotic manipulation tasks.

\end{itemize}

To assess the effectiveness of our proposed image editing framework in generating semantically accurate and visually consistent goal images, we conduct systematic comparisons against two representative methods with goal image generation: 
\begin{itemize}
\item \textbf{SUSIE} \cite{blackzero}: A two-stage instruction-driven framework that first generates goal images based on task prompts via image editing techniques, and then uses a goal-conditioned policy to execute actions. We follow the official evaluation protocol by using their released checkpoint directly and apply it to the BridgeDataset V2 benchmark.
\item \textbf{GHIL‑Glue} \cite{hatch2024ghil}: A hierarchical control algorithm designed to connect high-level generative models with low-level robot policies. It improves robustness by filtering out subgoal images that are visually inconsistent or do not contribute to task progression, and further enhances reliability through augmentation de‑synchronization strategies. 
\end{itemize}

\begin{figure*}[t]
  \centering
  \includegraphics[width=\textwidth]{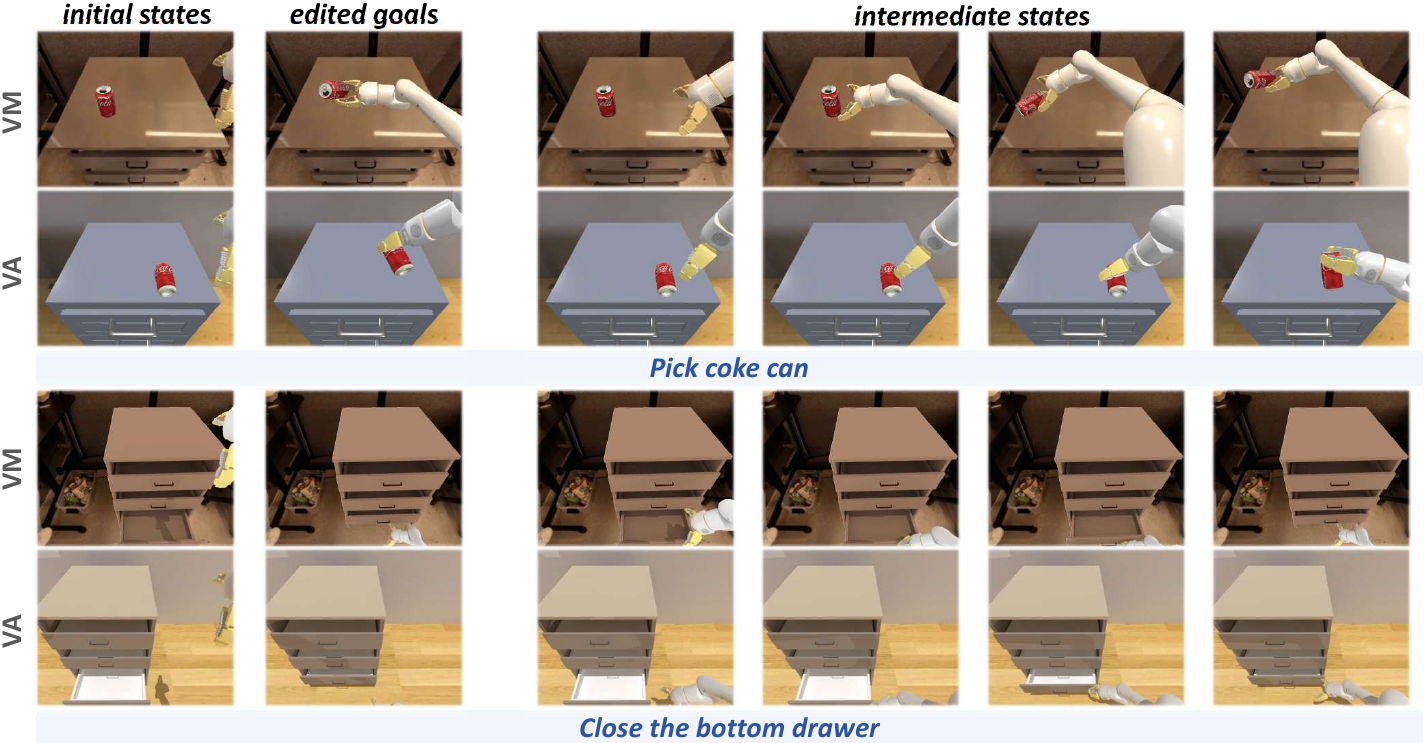}   
  \includegraphics[width=\textwidth]{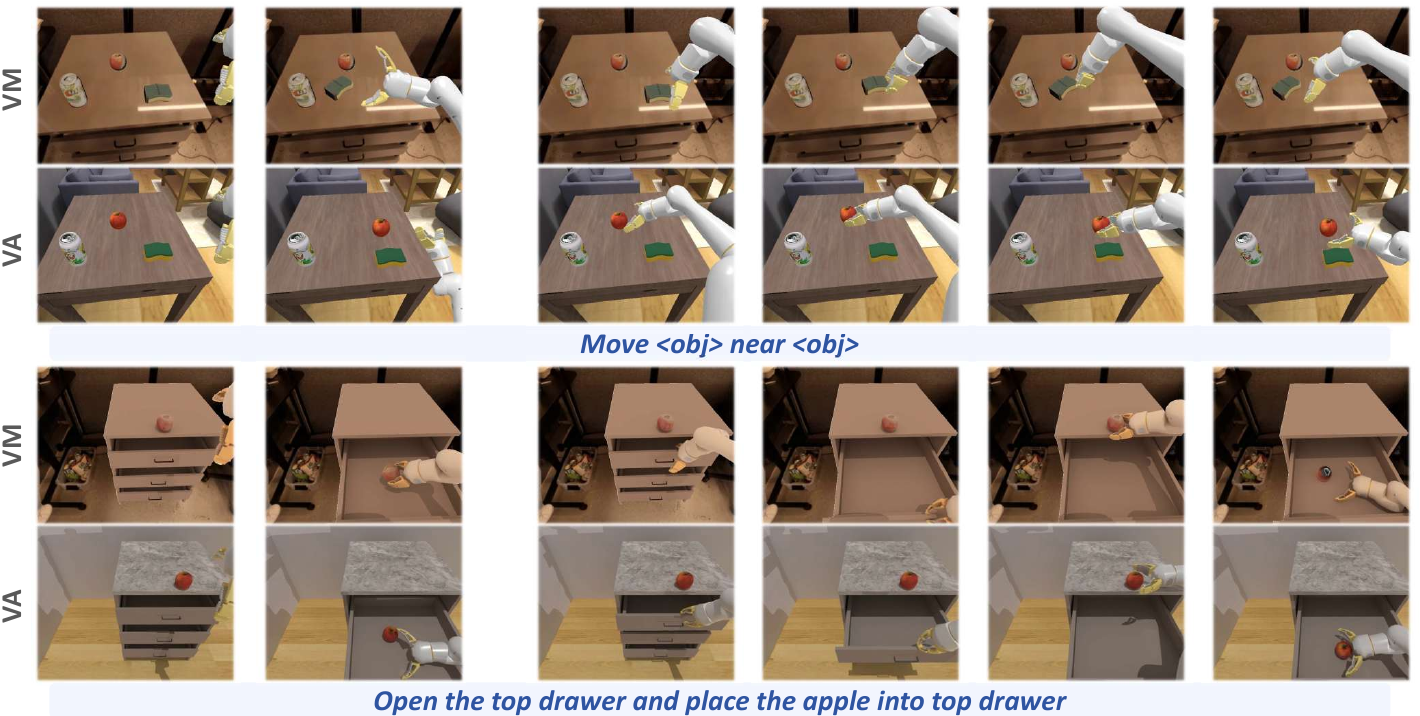}   
\caption{Qualitative results on the SIMPLER simulation environment using the Google Robot. Each row corresponds to a manipulation task evaluated under both Visual Matching (VM) and Variant Aggregation (VA) settings. For each task, we visualize the initial observation, the goal image edited and intermediate execution frames. 
}
  \label{fig11}
  \vspace{-8pt}
\end{figure*}



\section{Additional Analysis}
\subsection{Is Language Still Necessary?}
We further investigate whether goal images can substitute for language instructions during GIFT adaptation. Table~\ref{tab:language_ablation} compares CogACT, our proposed method, and a variant that removes language instructions while retaining the visual goal condition. The no-language variant collapses to 0\% success on both Grasp and Stack and exhibits unstable fine-tuning, whereas GIFT achieves clear gains over CogACT. These results show that goal images do not replace the pretrained language-conditioned policy structure. Consistent with our observation that GIFT acts as an efficient capability amplifier rather than a source of entirely new manipulation skills, language remains essential for preserving the pretrained multimodal reasoning prior, while visual goals refine task alignment within the existing behavioral scope of the original policy.
\begin{table}[t]
\centering
\caption{Input-modality ablation on representative SIMPLER tasks. Removing language instructions severely degrades performance even when goal images are provided, indicating that language remains essential for stable goal-aware adaptation.}
\small
\begin{tabular}{lcccc}
\toprule
Modalities        & Epochs    & Grasp      & Stack      & Stability \\
\midrule
V + L           & --        & 80.8       & 15.0       & -- \\
V + L + G       & 1         & 86.9       & 43.1       & Stable \\
V + G           & 1         & 0.0        & 0.0        & Unstable \\
\bottomrule
\end{tabular}
\label{tab:language_ablation}
  \vspace{-8pt}
\end{table}

\subsection{Robustness to Editing Stochasticity.}
Image editing models can introduce stochastic variations during goal-image generation, which may affect the downstream policy if the generated goals are unstable. To evaluate the robustness of GIFT to such editing stochasticity, we run the goal-generation process with three commonly used random seeds, 42, 22 and 62, on representative SIMPLER Visual Matching tasks. As shown in Table~\ref{tab:seed_variance}, GIFT exhibits stable performance across different seeds. The success rates vary only slightly on Grasp, ranging from 87.0\% to 88.3\%, and on Move, ranging from 85.7\% to 86.9\%. Drawer shows similarly small variation, with results between 62.9\% and 64.7\%. The standard deviation remains below 1\% for all three tasks, indicating that moderate stochasticity in generated goal images has limited impact on downstream policy execution under our evaluation setting.
\begin{table}[t]
\centering
\caption{Robustness of GIFT to the stochasticity of goal-image generation under three random seeds in the SIMPLER Visual Matching setting. The performance remains highly consistent across seeds, with standard deviations below 1\% on all tasks.}
\small
\begin{tabular}{lcccc}
\toprule
Task    & Seed 42 & Seed 22 & Seed 62 & Mean $\pm$ Std \\
\midrule
Grasp   & 87.9 & 88.3 & 87.0 & $87.7 \pm 0.5$ \\
Move    & 86.9 & 85.7 & 86.1 & $86.2 \pm 0.6$ \\
Drawer  & 64.7 & 62.9 & 63.0 & $63.5 \pm 0.8$ \\
\bottomrule
\end{tabular}
\vspace{-8pt}
\label{tab:seed_variance}
\end{table}

\subsection{Computational Overhead of GIFT}

\begin{table*}[t]
\centering
\caption{System-level computational overhead of GIFT across baseline policies. Profiling is conducted on a single NVIDIA L20 GPU in bf16 precision. Params'' and Step Time'' report base model costs plus GIFT's overhead. B / G'' denotes Base / GIFT. Edit'' is a one-time per-episode cost.}
\label{tab:overhead}
\resizebox{\textwidth}{!}{%
\begin{tabular}{lccccccccc}
\toprule
\textbf{Model} & \textbf{Params} & \textbf{Trainable} & \textbf{FLOPs} & \textbf{Mem} & \textbf{Step Time} & \textbf{Edit} & \textbf{Steps (B / G)} & \textbf{Ep. Cost (B / G)} & \textbf{Overhead} \\
\midrule
$\pi_{0.5}$ & 3B+20M   & 100M & 0.85T & 7.3G & 73ms+2ms & 4.2s & 52.4 / 50.8 & 6.45s / 10.55s & +63.7\% \\
CogACT      & 7B+72M   & 150M & 4.32T & 15.8G & 156ms+4ms & 4.2s & 85.3 / 77.5 & 14.16s / 17.38s & +22.7\% \\
OpenVLA     & 7B+72M   & 200M & 4.29T & 15.5G & 75ms+4ms & 4.2s & 125.5 / 113.6 & 10.67s / 14.31s & +34.1\% \\
SpatialVLA  & 4B+10M   & 630M & 1.94T & 8.5G & 97ms+4ms & 4.2s & 99.7 / 95.3 & 10.67s / 14.78s & +38.5\% \\
\bottomrule
\end{tabular}%
}
\end{table*}
To comprehensively evaluate the system-level efficiency of our framework, Table \ref{tab:overhead} summarizes both the parameter requirements and the runtime computational overhead introduced by GIFT across four baseline policies. Instead of only analyzing parameter counts, we also profile the FLOPs per step, peak inference memory, one-time goal-editing latency, per-step policy latency, and the end-to-end episode execution cost. All profiling is conducted on a single NVIDIA L20 GPU with bf16 precision.First, from an architectural perspective, GIFT introduces only a fraction of additional parameters relative to the underlying VLA models. The extra parameters range from 20M for $\pi_{0.5}$ to 72M for 7B-scale models (CogACT and OpenVLA). Consequently, GIFT maintains a strictly lightweight profile on the policy execution side. The ZeroConv injection mechanism adds only a marginal 2 to 4 ms of latency per step, and the increase in peak inference memory is minimal, which perfectly preserves the high-frequency control capabilities of the pretrained VLA models.From a system-level end-to-end perspective, the primary computational overhead originates from the one-time goal-image editing stage, which takes approximately 4.2 seconds prior to action execution. When factoring in this initialization stage along with the average rollout length and environment step time, GIFT introduces an overall episode-level overhead ranging from 22.7\% (CogACT) to 63.7\% ($\pi_{0.5}$).However, it is worth noting that explicit visual goal guidance often empowers the policy to complete manipulation tasks more decisively, resulting in fewer execution steps on average (e.g., OpenVLA average steps decrease from 125.5 to 113.6). Ultimately, while the current diffusion-based editor introduces a visible one-time latency at the beginning of an episode, the policy-side adaptation remains highly efficient. Future integration with accelerated single-step diffusion models (e.g., SwiftEdit or TurboEdit) could further minimize this end-to-end deployment overhead.


\section{Editing Example}
\begin{figure*}[t]
  \centering
  \includegraphics[width=\textwidth]{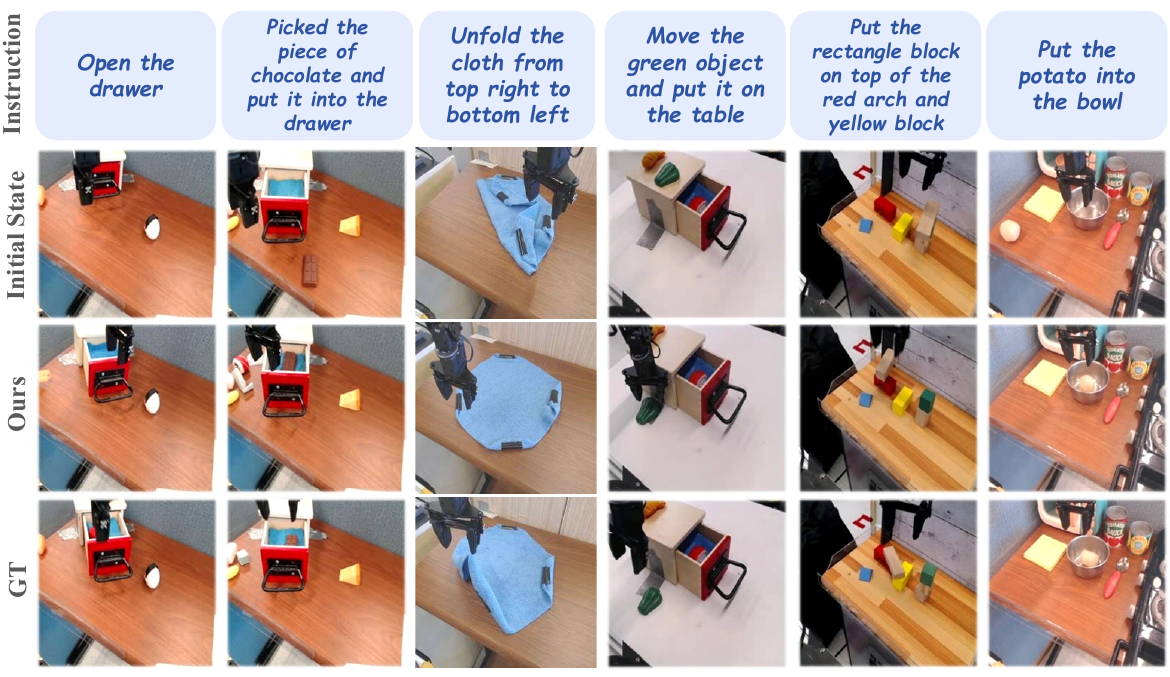}   
  \includegraphics[width=\textwidth]{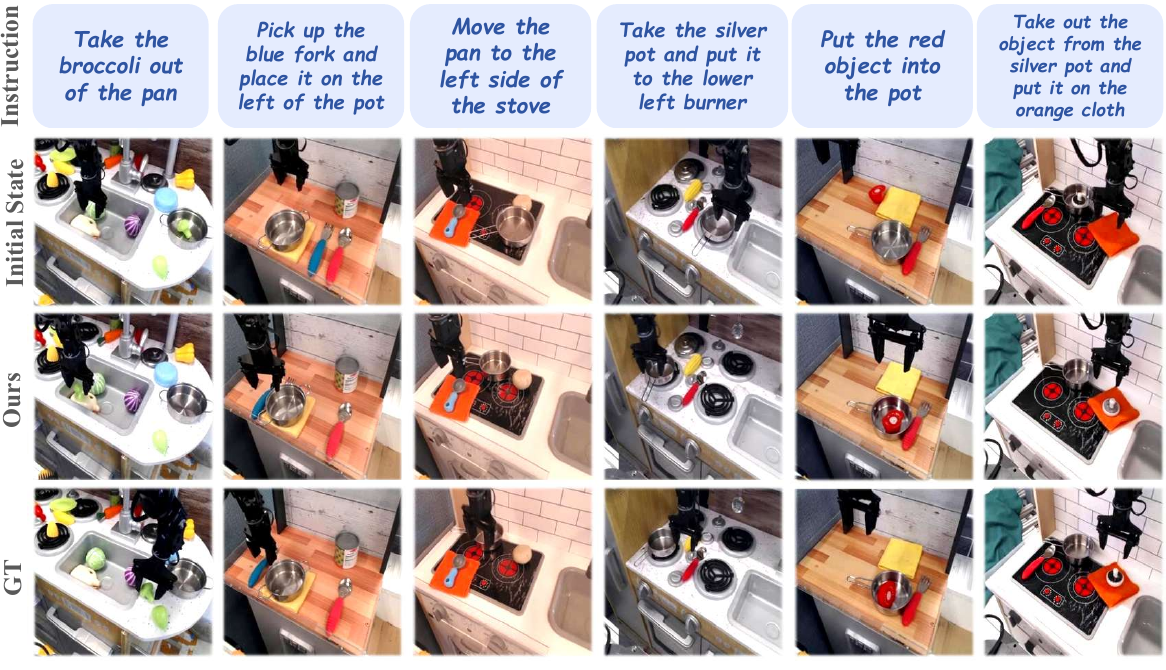}   
\caption{Qualitative results of edited goal images generated by our method (\textbf{Ours}) across diverse manipulation tasks from our test dataset. Each column represents a distinct task, with the initial state shown in the first row. Our framework successfully produces visually and semantically aligned goal states, preserving spatial and object-specific cues necessary for grounding manipulation actions, even under complex instructions and cluttered scenes.}

  \label{fig7}
  \vspace{-8pt}
\end{figure*}

To evaluate the visual quality and instruction fidelity of our image editing framework, we visualize generated goal images on held-out samples from our constructed test set. The test set is derived from the remaining 10\% of BridgeDataset V2, following training on a mixed dataset that includes 90\% of BridgeDataset V2 and 30\% of LIBERO. Each sample consists of an initial observation, a language instruction, and a target goal image. As shown in Figure~\ref{fig7}, our model produces visually realistic and semantically consistent results across a variety of manipulation goals, reflecting strong alignment with task intents and visual grounding in both tabletop and kitchen-like scenes.

\section{Visual Example}

To provide a more intuitive understanding of our model's effectiveness, we visualize representative qualitative results across both simulation and real-world manipulation settings. Specifically, we select diverse examples from two simulation environments, SIMPLER and LIBERO, covering different robot embodiments such as Google Robot and WidowX, as well as varied task configurations. In addition, we include real-robot rollouts under both in-domain and out-of-distribution lighting conditions. These visualizations show how our framework edits goal images and guides policy execution across different task complexities, object categories, robot platforms, and visual conditions, demonstrating its effectiveness in both simulated benchmarks and real-world deployment.

\subsection{Qualitative Results on SIMPLER}

\paragraph{Google Robot.} 
Figure~\ref{fig11} showcases the performance of our method across a diverse set of complex manipulation tasks using the Google Robot in the SIMPLER simulation environment. Under both the Visual Matching and Variant Aggregation evaluation protocols, our framework consistently produces task-relevant goal images that maintain spatial coherence and semantic fidelity, even in visually diverse and dynamically changing scenes. These edited goal representations allow the policy to remain closely aligned with the intended task objectives throughout execution, demonstrating strong robustness to visual variation and scene complexity. Furthermore, the coherent execution progress reflected in intermediate states provides additional evidence of the effectiveness of goal-conditioned guidance.

\paragraph{WidowX.}
Figure~\ref{fig10} presents qualitative results from the SIMPLER benchmark using the WidowX robot. These examples demonstrate the ability of our image editing model to support a broad range of fine-grained object manipulation tasks. Despite substantial variations in object geometry, spatial placement, and background appearance, our framework consistently generates goal images that are both visually plausible and closely aligned with the given instructions. Overall, these results underscore the robustness of our goal-conditioned guidance mechanism, particularly in visually cluttered environments and instruction-sensitive manipulation scenarios.

\subsection{Qualitative Results on LIBERO.}
Figure~\ref{fig9} demonstrates our model’s strong generalization capability across a broad spectrum of manipulation scenarios in the LIBERO benchmark. Despite substantial variations in object categories, spatial arrangements, and goal semantics across different tasks, our approach consistently produces goal images that are both visually accurate and closely aligned with the provided instructions. These edited goal representations effectively ground a range of complex manipulation behaviors, including container placement, drawer interaction, and coordinated multi-object manipulation. Moreover, the intermediate execution states indicate that goal image guidance plays a critical role in shaping the robot’s action trajectories over time, facilitating successful task execution that requires sequential reasoning and precise spatial planning. Notably, the model maintains high visual and semantic fidelity across both simple and compound goals, further highlighting its robustness, adaptability, and reliability in challenging simulation environments.

\subsection{Qualitative Real-Robot Evaluation}
Figure~\ref{fig:real_robot_rollouts} presents qualitative real-robot executions under both in-domain (ID) and out-of-distribution (OOD) lighting conditions. The first task requires the robot to pick up the bottle, while the second task requires placing the bottle into the basket. For each task, we visualize the full execution sequence from the third-view camera. Under ID lighting, the robot accurately approaches the target object and completes the manipulation. Under OOD lighting, although the visual appearance of the scene changes noticeably, the robot remains able to identify the task-relevant object, execute the grasp, and complete the goal-directed behavior. These rollouts demonstrate that the GIFT-enhanced policy can preserve robust real-world execution under visual distribution shifts, supporting both object-centric grasping and goal-directed placement.
\end{document}